\documentclass[sigconf]{aamas}  % do not change this line!

\usepackage{booktabs}
\usepackage[linesnumbered, ruled, vlined]{algorithm2e}
\usepackage{hyperref}
\setcopyright{ifaamas}  % do not change this line!
\acmDOI{}  % do not change this line!
\acmISBN{}  % do not change this line!
\acmConference[AAMAS'18]{Proc.\@ of the 17th International Conference on Autonomous Agents and Multiagent Systems (AAMAS 2018)}{July 10--15, 2018}{Stockholm, Sweden}{M.~Dastani, G.~Sukthankar, E.~Andr\'{e}, S.~Koenig (eds.)}  % do not change this line!
\acmYear{2018}  % do not change this line!
\copyrightyear{2018}  % do not change this line!
\acmPrice{}  % do not change this line!

\begin{document}

\title{Two Techniques That Enhance the Performance of Multi-robot Prioritized Path Planning}  % put your title here!
%\titlenote{Produces the permission block, and copyright information}

% AAMAS: as appropriate, uncomment one subtitle line; see camera ready instructions
\subtitle{Extended Abstract}
%\subtitle{Industrial Applications Track}
%\subtitle{Socially Interactive Agents Track}
%\subtitle{Blue Sky Ideas Track}
%\subtitle{Robotics Track}
%\subtitle{JAAMAS Track}
%\subtitle{Doctoral Mentoring Program}

%\subtitlenote{The full version of the author's guide is available as \texttt{acmart.pdf} document}

% replace this with your author block!
\author{Anton Andreychuk}
\affiliation{%
  \institution{Peoples' Friendship University of Russia\\(RUDN University)}
  \city{Moscow} 
  \state{Russia} 
}
\email{andreychuk@mail.com}

\author{Konstantin Yakovlev}
\affiliation{%
  \institution{Federal Research Center ``Computer Science and Control'' of Russian Academy of Sciences}
%  \institution{National Research University Higher School of Economics}
%  \city{Moscow} 
%  \state{Russia} 
}
\affiliation{%
%  \institution{Federal Research Center ``Computer Science and Control'' of Russian Academy of Sciences}
  \institution{National Research University Higher School of Economics}
  \city{Moscow} 
  \state{Russia} 
}
\email{yakovlev@isa.ru}

\begin{abstract}  % put your abstract here!
We introduce and empirically evaluate two techniques aimed at enhancing the performance of multi-robot prioritized path planning. The first technique is the deterministic procedure for re-scheduling (as opposed to well-known approach based on random restarts), the second one is the heuristic procedure that modifies the search-space of the individual planner involved in the prioritized path finding.

\end{abstract}

  % put your semicolon-separated keywords here!

\maketitle

%%%%%%%%%%%%%%%%%%%%%%%%%%%%%%%%%%%%%%%%%%%%%%%%%%%%%%%%%%%%%%%%%%%%%%%%%%%%%%%%%%%%%%%%%%%%%%%%%%%%%%%%%
%% start of main body of paper

\section{Problem statement}\label{problem}
Consider $n$ open-disk robots of equal radii $r$ intended to simultaneously move towards their goals on a grid comprised of blocked and unblocked cells. The task is to plan the set of feasible trajectories (one per each robot) such that each two trajectories for different robots are collision-free, e.g. robots following them never collide. The spatial component of the trajectory, e.g. the path, is the sequence of straight-line segments connecting the centers of grid cells. We assume that all robots follow their paths with identical speed and can start/stop instantaneously. Depending on the application domain the cost of the solution can be either \textit{makespan}, i.e. the time by which the last robot reaches its goal, or \textit{flowtime}, which is the sum of traversal times across all the robots involved in the instance.

\section{Method}\label{method}
\subsection{Related work}
Complete (and optimal or bounded-suboptimal) solvers to the described problem exist, like the ones introduced in \cite{yu2016optimal}, \cite{standley2010}, \cite{sharon2015}, \cite{Wagner2011}, \cite{cohen2016}, \cite{ma2017lifelong} etc., but they typically require significant computational effort, especially when the number of robots increases. Appealing alternative is prioritized planning \cite{erdmann1987}, which is widely used in robotics \cite{clark2001randomized}, \cite{cap2015a} due to its simplicity and ability to scale well to large problems. Prioritized planners are incomplete in general but can be tweaked to increase the chance of finding the solution (and decrease flowtime and/or makespan). The most widespread approach to do so is to randomly re-assign priorities in case of failure and re-plan, see \cite{bennewitz2001optimizing}, \cite{bennewitz2002finding} for details.

\subsection{Contribution}
In this work we introduce a novel deterministic technique for re-assigning the priorities (re-scheduling) in case planning with initial priorities fails. We also introduce the notion of safe-start-interval (\textit{SSI}) in order to prohibit the individual planner, involved in prioritized planning, from interfering with the start positions of low-priority robots for certain amount of time. Both suggested techniques enhance the performance of  planning (as shown experimentally) and can be embedded into any prioritized planner.

\subsection{Deterministic re-scheduling}
Prioritized planners are incremental in nature, e.g. they plan individual trajectories sequentially one by one in accordance with the imposed priority ordering. The planner may fail to solve an instance only by failing to find an individual trajectory for some robot in a sequence, which, in turn, happens only when the high-priority robots prevent it from doing so by their motion\footnote{The case when no path exists due to static obstacles is not considered in this work as being out of interest.}. Thus, in case of failure, we suggest raising the priority of the failed robot to 1 (the highest priority) and re-planning. Obviously this robot will not be the cause of multi-agent planner's failure anymore (which is not true with random re-scheduling) and chances are the overall solution will be found. If no, deterministic re-scheduling process repeats until the planner succeeds or until it comes across the priority sequence that has already been investigated (in this case "failure" is reported).

\subsection{Start-safe intervals}
In conventional prioritized planning high priority robots are allowed to pass through the start locations of the low-priority ones at any moment of time. This might be undesirable as further on some robot may fail to plan its trajectory due to the fact it's immediately knocked off by the high-priority robot and there is no room for him to step away. To mitigate this issue we introduce the notion of \textit{safe-start-interval} (\textit{SSI}). \textit{SSI} is a time interval $[0, k], k\geq0$, during which any robot is prohibited to interfere with the start location of any other robot (and it is allowed afterwards). This introduces flexibility that can positively influence planner's effectiveness, e.g. the ability to solve "unsolvable" instances (conducted experimental evaluation supports this claim).

\section{Experimental evaluation}
To evaluate the proposed techniques and to estimate how they influence the performance of prioritized planning we experimented with two different types of environments, e.g. the one with no obstacles present (32 x 32 empty grid) and the one modeling the warehouse from \cite{ma2017lifelong} (21x35 grid with 100 blocked cells out of 735). The number of agents differed from 8 to 320 depending on the environment. 100 instances per each number of agents per environment were randomly generated and used for evaluation. Initial scheduling was "shortest first", which means that robots with the lower estimates of path lengths are assigned higher priorities. All experiments were conducted using C++ implementation of the state-of-the-art prioritized planner AA-SIPP(m) \cite{yakovlev2017aasipp}\footnote{Sources are available at \href{https://github.com/PathPlanning/AA-SIPP-m}{https://github.com/PathPlanning/AA-SIPP-m}}.

We will now highlight only a small subset of experimental results that provide a strong support for the claim that introducing \textit{SSI} and deterministic re-scheduling positively influences the performance of prioritized multi-robot path planning compared to baseline (no re-scheduling, no safe-start-intervals) and to random re-scheduling.  

\begin{figure}[t]
\includegraphics[width=\columnwidth]{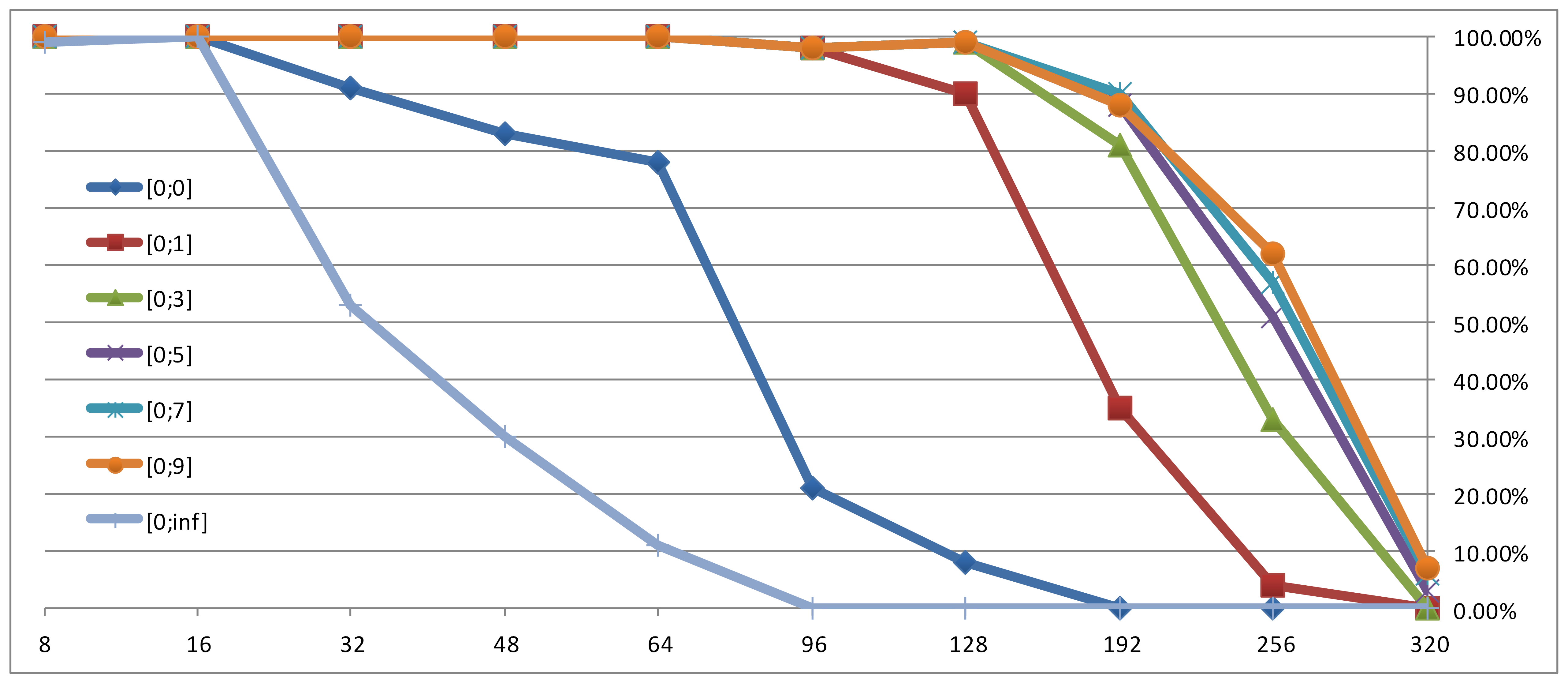}
\caption{Success rate (y-axis) of the prioritized planner with different safe-start-intervals for 8-320 agents (x-axis) on empty 32x32 grid.}
\label{fig1}
\end{figure}

\figurename \ \ref{fig1} depicts the success rate of the planner on 32x32 empty grid with no re-scheduling and \textit{SSI} endpoints set to 0 (baseline, no \textit{SSI}), 1, 3, 5, 7, 9 and  $\infty$ (traversing the start locations of the low-priority robots is completely prohibited). As one can see introducing \textit{SSI} leads to a significant increase of the success rate, especially when the number of robots is large. For example, more than 80\% of instances for 192 agents (density 18.75\%) are solved when \textit{SSI} endpoint is set to $k\geq3$ compared to 0\% for planning without \textit{SSI}. Setting $k$ to infinity is the worst option, although it is known from previous research \cite{cap2015a} that such setting guarantees finding a solution in case given path planning instance satisfies certain criteria. 

Table \ref{tab1} shows the success rate, runtime and quality indicators (makespan, flowtime) of the planner relying on random re-prioritization and on the suggested deterministic re-prioritization for warehouse-like environment from \cite{ma2017lifelong}. \textit{SSI} was set to 5 and 5 minutes time cap was introduced for this experiment as the planner relying on random re-scheduling might run for inadequately long times (as number of restarts is not limited). Randomized planner was invoked 10 times on each instance and all the collected data was used for averaging.  

\begin{table}[t]
  \caption{Success rate (SR) and average runtime (t), makespan (Msp) and flowtime (Flt) of the prioritized planner with random and deterministic re-scheduling for 16-160 agents (warehouse environment).}
  \label{tab1}
  \resizebox{\columnwidth}{!}{\begin{tabular}{|c|cccc|cccc|}
   \hline
   &\multicolumn{4}{|c|}{Random re-scheduling}&\multicolumn{4}{|c|}{Deterministic re-scheduling}\\
   &SR&t&Msp&Flt&SR&t&Msp&Flt\\
   \hline
    16&1.00&0.01&33.31&285.15&1.00&0.01&33.31&285.15\\
    32&1.00&0.04&36.68&591.30&1.00&0.04&36.68&591.30\\
    64&1.00&0.22&41.25&1274.94&1.00&0.17&42.05&1252.82\\
    96&1.00&0.68&50.05&2212.14&1.00&0.49&50.42&2149.31\\
    128&1.00&5.73&62.23&3543.92&1.00&1.40&66.15&3322.52\\
    160&0.42&117.82&79.73&5220.90&0.99&28.28&100.50&5298.85\\
    192&0.00& --- & --- & --- &0.00& --- & --- & --- \\
  \bottomrule
\end{tabular}}
\end{table}

As one can see the success rate tends to instantaneous drop-down after some critical density of agents is reached, but the proposed deterministic re-scheduling technique leads to better success rate, when this density is near-critical (160 robots in the considered case). Moreover, utilizing the suggested approach results in slightly better flowtime. Finally, the proposed technique is much faster compared to random re-scheduling. The difference in runtimes is evident for higher densities when the need for attempting different priority orderings is vital. For example for 160 agents (density 25\%) our approach, besides solving almost 100\% of tasks, compared to 42\% achieved by random restarts, produced solutions 4 times faster. The only downside of the suggested deterministic re-scheduling procedure is the increase of makespan. In case it is of extreme importance one might consider to change the initial priority scheduling scheme to "longest first" as suggested in \cite{van2005prioritized} and to apply randomized restarts. 

\section{Conclusions}
In this work we have studied the behavior of prioritized multi-robot path finders applied to the problem of navigating disk-shaped robots on uniform cost grids. We have suggested two novel procedures that can be implemented on top of any prioritized solver and evaluate them empirically. The experiments demonstrate that they drastically reduce the number of failures when compared to baseline algorithm. At the same time, the planner utilizing them outperforms the one relying on random restarts (the most commonly used re-scheduling technique in prioritized path planning) both in terms of effectiveness, e.g. success rate, and efficiency, e.g. runtime.

\begin{acks}
    This work was supported by the Russian Science Foundation (Project No. 16-11-00048).

\end{acks}

%%%%%%%%%%%%%%%%%%%%%%%%%%%%%%%%%%%%%%%%%%%%%%%%%%%%%%%%%%%%%%%%%%%%%%%%%%%%%%%%%%%%%%%%%%%%%%%%%%%%%%%%%
%% bibliography: see CFP for number of permitted pages

\bibliographystyle{ACM-Reference-Format}  % do not change this line!
\bibliography{MAPF}  % put name of your .bib file here

\end{document}